\documentclass[10pt,twocolumn,letterpaper]{article}

\usepackage{cvpr}      %

\usepackage[dvipsnames]{xcolor}

\usepackage{graphicx}
\usepackage{booktabs}
\usepackage{xcolor,colortbl}
\usepackage{caption}
\usepackage{graphicx}
\usepackage{tikz}
\usepackage{subcaption}
\usepackage{amsmath}
\usepackage[bb=dsserif]{mathalpha}
\usepackage{bm}
\usepackage{floatrow}
\usepackage{float}
\definecolor{cvprblue}{rgb}{0.21,0.49,0.74}
\usepackage[pagebackref,breaklinks,colorlinks,citecolor=cvprblue,linkcolor=cvprblue,urlcolor=cvprblue]{hyperref}

\def\paperID{00007} %
\def\confName{3DV\xspace}
\def\confYear{2025\xspace}

\title{AugSplat: Radiance Field-Informed Gaussian Splatting for Sparse-View Settings}

\author{
Lorenzo Lazzaroni$^{2}$ \hspace{.5cm}
Riccardo Bollati$^{2}$ \hspace{.5cm}
Daniel Barath$^{1,2}$ \\
Michael Niemeyer$^{1}$ \hspace{.5cm}
Keisuke Tateno$^{1}$ \\
$^{1}$Google \hspace{.5cm} $^{2}$ETH Zurich \\
{\tt\small \href{https://github.com/llazzaroni/AugSplat}{github.com/llazzaroni/AugSplat}}
}

\makeatletter
\DeclareRobustCommand\onedot{\futurelet\@let@token\@onedot}
\def\@onedot{\ifx\@let@token.\else.\null\fi\xspace}

\makeatother

\newcommand{\boldparagraph}[1]{\vspace{0.2cm}\noindent{\bf #1.}}

\begin{document}
\maketitle
\begin{abstract}
  Generating high-quality novel views at real-time frame rates remains a central challenge in 3D vision, particularly in sparse-view scenarios. Neural radiance fields have demonstrated robust reconstruction from limited observations, but their reliance on volumetric rendering leads to high computational cost and slow inference. In contrast, Gaussian Splatting methods achieve real-time rendering through rasterization, but their optimization is highly sensitive to the quality of the initial geometry. This sensitivity becomes especially problematic in sparse-view settings, where limited observations often lead to incomplete or noisy point-cloud reconstructions. In this work, we present \textit{AugSplat}, a simple framework for improving Gaussian Splatting in sparse-view regimes using radiance-field-based view augmentation. We first train a radiance field on the sparse input views and use it to synthesize additional images from nearby novel viewpoints, increasing the effective view-space coverage available for supervision. These synthetic views are then used as auxiliary supervision during Gaussian Splatting optimization. We study two variants: \textit{Staged AugSplat}, which uses synthetic views for an initial optimization phase before switching to real images, and \textit{Dual AugSplat}, which jointly trains on real and synthetic views with a decaying synthetic loss weight. Experiments on sparse-view mip-NeRF 360 scenes show that AugSplat improves reconstruction quality over standard Gaussian Splatting. Staged AugSplat achieves the strongest average performance, while Dual AugSplat provides a closely performing formulation that keeps real-image supervision active throughout training, and both variants preserve real-time rendering at inference.
\end{abstract}    
\section{Introduction}
\label{sec:intro}

High-quality novel view synthesis at real-time frame rates is a central objective in 3D vision, with applications ranging from virtual reality to robotics. Neural radiance fields (NeRFs)~\cite{mildenhall2020nerf} and related neural field representations~\cite{mescheder2019occupancynet, park2019deepsdf, chen2018implicit_decoder,xie2022neural} have driven substantial progress in this area by representing scenes as continuous functions and rendering images through volumetric integration. Due to their continuous formulation and robust optimization behavior, neural field methods have demonstrated strong performance in challenging real-world settings~\cite{brualla2021wild}. However, their reliance on volumetric rendering often leads to high computational cost, limiting their applicability when real-time inference is required.

Recently, 3D Gaussian Splatting (3DGS)~\cite{kerbl20233d} has emerged as an alternative paradigm, providing an efficient alternative thanks to its rasterization-based nature. By representing scenes as collections of anisotropic Gaussian primitives, 3DGS achieves high-quality rendering at real-time frame rates. Despite this advantage, its optimization is highly dependent on the quality of the initial geometry. This limitation is particularly pronounced in sparse-view settings, where limited image observations often lead to incomplete or noisy point-cloud reconstructions. In such cases, Gaussian primitives may be poorly placed or missing in under-observed regions, making it difficult for optimization to recover accurate geometry and appearance.

Sparse-view reconstruction~\cite{Niemeyer2021Regnerf,Jain_2021_ICCV,yu2021pixelnerf} remains especially challenging in this context. With only a small number of input views, traditional structure-from-motion pipelines~\cite{Schonberger_2016_CVPR,wang2023visualgeometrygroundeddeep} may fail to recover sufficient scene structure, which directly affects downstream Gaussian Splatting optimization. While neural radiance fields can often produce plausible reconstructions from sparse observations due to their continuous volumetric representation, Gaussian Splatting typically lacks an equally flexible mechanism for reasoning about regions that are weakly supported by the initial point cloud.

In this work, we propose \textit{AugSplat}, a simple framework for improving Gaussian Splatting in sparse-view regimes using radiance-field-based view augmentation. Our key idea is to use a radiance field not as the final rendering representation, but as a source of additional supervision for Gaussian Splatting. Given sparse input views, we first train a radiance field and use it to synthesize images from nearby novel viewpoints. These synthetic views increase the effective supervisory coverage of the sparse input set and provide additional constraints during Gaussian optimization.

RadSplat~\cite{niemeyer2025radsplat} is closely related to our work, as it also leverages radiance fields to improve Gaussian Splatting. However, while RadSplat uses radiance-field information to improve Gaussian initialization, AugSplat uses radiance-field renderings as auxiliary training views. This makes our method complementary to initialization-based approaches.

We explore two variants of this strategy. \textit{Staged AugSplat} first optimizes Gaussian Splatting using synthetic radiance-field views before switching to real images, while \textit{Dual AugSplat} jointly trains on real and synthetic views with a decaying weight on the synthetic supervision. By using radiance fields to augment sparse supervision and Gaussian Splatting for efficient inference, AugSplat combines the robustness of implicit representations with the real-time rendering speed of explicit Gaussian primitives.

In summary, our contributions are as follows:
\begin{itemize}
    \item We introduce \textit{AugSplat}, a radiance-field-guided augmentation framework for improving Gaussian Splatting in sparse-view settings.
    \item We propose two training strategies, \textit{Staged AugSplat} and \textit{Dual AugSplat}, for incorporating synthetic radiance-field views into Gaussian Splatting optimization.
    \item We show that radiance-field-based view augmentation improves sparse-view Gaussian Splatting on Mip-NeRF360~\cite{barron2022mip} scenes. Staged AugSplat achieves the strongest average reconstruction quality, while Dual AugSplat provides a closely performing formulation that keeps real-image supervision active throughout training.
    \item We analyze training dynamics and show that AugSplat improves early-stage optimization rather than merely benefiting from longer training.
\end{itemize}
\section{Related Work}
\label{sec:relwork}

\boldparagraph{Neural Radiance Fields}
Neural radiance fields (NeRFs)~\cite{mildenhall2020nerf} represent scenes as continuous volumetric functions, enabling high-quality novel view synthesis by modeling both geometry and appearance. Subsequent work has improved reconstruction quality, scalability, and anti-aliasing for complex scenes~\cite{barron2022mip,Barron2023zipnerf}, while efficient neural representations such as Instant-NGP~\cite{mueller2022instant} have substantially reduced training and rendering costs. Nevertheless, many NeRF-based methods still rely on volumetric rendering and therefore do not generally provide the same real-time inference performance as rasterization-based Gaussian Splatting.

In AugSplat, we use this robustness not for final rendering, but to synthesize additional supervision for a real-time Gaussian representation.

\boldparagraph{Sparse-View Novel View Synthesis}
Novel view synthesis from sparse observations is highly under-constrained, since large portions of the scene may be weakly observed or entirely unseen. NeRF-based methods address this problem using regularization, learned priors, or auxiliary geometric constraints. For example, RegNeRF~\cite{Niemeyer2021Regnerf} regularizes geometry and appearance in unobserved views, while SparseNeRF~\cite{wang2023sparsenerf} leverages depth-ranking constraints to improve few-shot reconstruction.

More recently, sparse-view settings have also been studied for Gaussian Splatting. Methods such as SparseGS~\cite{xiong2025sparsegs} and DNGaussian~\cite{Li_2024_CVPR} introduce additional constraints, depth priors, or regularization strategies to improve Gaussian optimization from limited views. Our work is complementary to these approaches: rather than directly regularizing the Gaussian representation, we use a radiance field to generate additional synthetic training views that increase the effective coverage of the sparse input set.

\boldparagraph{Gaussian Splatting and Extensions}
3D Gaussian Splatting (3DGS)~\cite{kerbl20233d} represents scenes using anisotropic Gaussian primitives and renders them efficiently through rasterization. This formulation enables high-quality real-time novel view synthesis, but also makes optimization sensitive to the quality of the initial point cloud and the distribution of training views. This sensitivity becomes particularly problematic in sparse-view settings, where incomplete geometry can lead to missing or poorly placed Gaussian primitives.

A large body of work has extended Gaussian Splatting to applications such as human and avatar reconstruction~\cite{zheng2024gpsgaussian,dhamo2023headgas,qian2024gaussianavatars,moreau2024human}, 3D generation~\cite{tang2024dreamgaussian,yi2024gaussiandreamer,chung2023luciddreamer}, SLAM~\cite{yan2023gs,li2024sgs,matsuki2024gaussian,Sandstrom_2025_CVPR}, dynamic scene reconstruction~\cite{wu20244dgaussians,xu2024fourkfourd}, and open-vocabulary scene understanding~\cite{shi2024language,qin2024langsplat}. These works demonstrate the flexibility of Gaussian representations, while our focus is specifically on improving their robustness under sparse-view supervision.

\boldparagraph{Combining Neural Fields and Gaussian Splatting}
Several recent methods combine neural field and Gaussian Splatting representations to exploit their complementary strengths. RadSplat~\cite{niemeyer2025radsplat} is closest to our work, as it also uses a radiance field to improve the robustness of Gaussian Splatting. Rather than deriving Gaussian primitives or representation-level priors from the radiance field, AugSplat treats the radiance field as a data generator: it renders nearby synthetic views and uses ensemble uncertainty to weight their contribution during optimization.

Other approaches integrate neural components more directly into Gaussian representations~\cite{malarz2024gaussiansplattingnerfbasedcolor,mihajlovic2024splatfieldsneuralgaussiansplats}, or use learned structures to predict Gaussian attributes in a view-adaptive manner~\cite{lu2023scaffoldgsstructured3dgaussians}. In contrast, AugSplat keeps the Gaussian representation unchanged and instead uses radiance-field renderings as additional training views.
\section{Method}
\label{sec:method}

Given a sparse set of calibrated input images, AugSplat first trains a radiance-field ensemble to obtain a continuous scene representation and uncertainty estimates. We then render nearby synthetic views from this ensemble and use the resulting images and confidence maps as auxiliary supervision for Gaussian Splatting. We consider two ways of incorporating this supervision: a staged strategy that uses synthetic views only during an initial warm-up phase, and a dual strategy that jointly optimizes real and synthetic supervision with a decaying synthetic loss weight.

\subsection{Neural Radiance Fields for Improved Supervisory Coverage}

A Neural Radiance Field (NeRF) represents a function
\[
f_\phi: \mathbb{R}^3 \times \mathbb{S}^2 \rightarrow \mathbb{R}^+ \times \mathbb{R}^3,
\]
mapping a 3D point $\mathbf{x}$ and viewing direction $\mathbf{d}$ to a volume density $\sigma$ and color $\mathbf{c}$. Given a camera ray, pixel colors are obtained by alpha compositing samples along the ray:
\begin{equation}
    \mathbf{c}_{\text{NeRF}} = \sum_{k=1}^{K} \tau_k \alpha_k \mathbf{c}_k,
\end{equation}
where
\begin{equation}
    \tau_k = \prod_{j=1}^{k-1} (1-\alpha_j), \quad
    \alpha_k = 1 - e^{-\sigma_k \delta_k}.
\end{equation}

The radiance field parameters $\phi$ are optimized by minimizing the photometric reconstruction loss over sampled training rays:
\begin{equation}
    \mathcal{L}_{\text{NeRF}}(\phi) =
    \sum_{\mathbf{r} \in \mathcal{R}_{\text{batch}}}
    \| \mathbf{c}^{\phi}_{\text{NeRF}}(\mathbf{r})
    - \mathbf{c}_{\text{image}}(\mathbf{r}) \|_2^2.
\end{equation}

\boldparagraph{Synthetic View Augmentation}
We leverage neural radiance fields to increase the effective geometric coverage available during Gaussian Splatting optimization. Given a sparse set of input views, we train an ensemble of $M$ radiance fields. The ensemble provides both synthetic color predictions and an uncertainty estimate, allowing us to downweight synthetic supervision in regions where the radiance field predictions disagree. We use this confidence weighting as a practical heuristic to reduce the influence of uncertain synthetic pixels.

We then sample a set of nearby novel camera poses
$\mathcal{C}_{\text{syn}} = \{ \mathbf{T}_i \}_{i=1}^{N_{\text{syn}}}$
by interpolating between existing training poses and applying small pose perturbations. We restrict these poses to remain close to the observed camera distribution, which helps maintain reliable radiance-field renderings while expanding scene coverage.

\begin{figure*}[t]
\centering

\begin{tikzpicture}
\node[anchor=south west,inner sep=0] (img) at (0,0)
{\includegraphics[width=\textwidth,trim={0 1.5cm 0 0},clip]{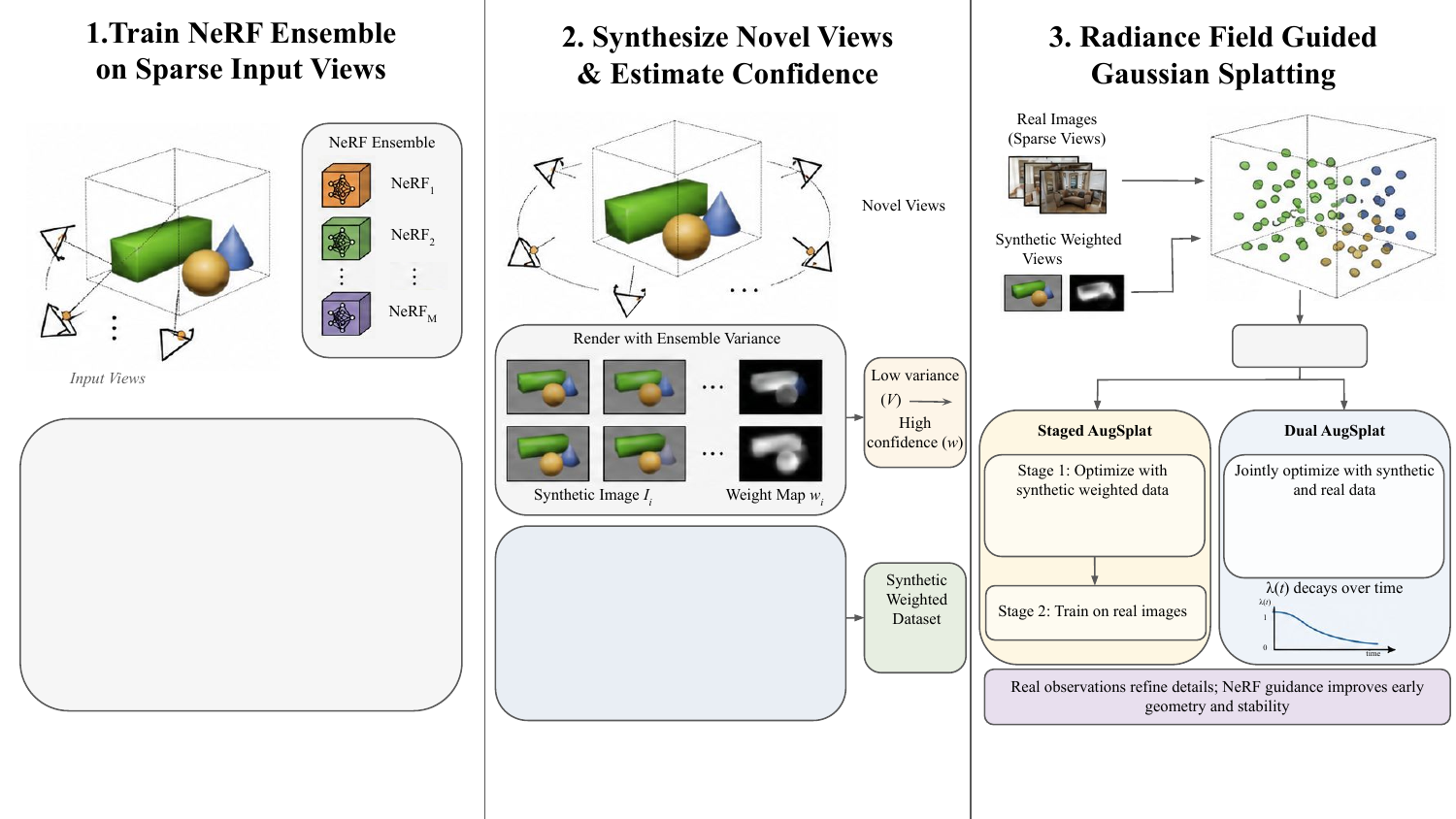}};

\begin{scope}[
x={(img.south east)},
y={(img.north west)}
]

% First formula box
\node[align=center,scale=0.9,text width=0.4\textwidth]
at (0.17,0.23)
{
\footnotesize
$\displaystyle f:\mathbb{R}^3\times S^2\rightarrow \mathbb{R}_+\times\mathbb{R}^3$

\vspace{0.2em}

$\displaystyle
\mathbf c_{\text{NeRF}}
=
\sum_{i=1}^{N}\tau_i\alpha_i\mathbf c_i
$

\vspace{0.2em}

$\displaystyle
\tau_i
=
\prod_{j=1}^{i-1}(1-\alpha_j),
\qquad
\alpha_i
=
1-e^{-\sigma_i\delta_i}
$

\vspace{0.2em}

$\displaystyle
\mathcal L(\phi)
=
\sum_{r\in\mathcal R_{\text{batch}}}
\left\|
c^{\phi}_{\text{NeRF}}(r)-c_{\text{image}}(r)
\right\|_2^2
$
};

% Second formula box
\node[align=center,scale=0.85,text width=0.28\textwidth]
at (0.46,0.153)
{
\footnotesize
$\displaystyle
I_i
=
\operatorname{median}_{j=1}^{M}
I_j^{\text{nerf}}(T_i)
$

\vspace{0.4em}

$\displaystyle
V_{\text{norm}}(x,y)
=
\operatorname{clip}
\!\left(
\frac{V(x,y)-lo}{hi-lo},
0,1
\right)
$

\vspace{0.4em}

$\displaystyle
w(x,y)
=
(1-V_{\text{norm}}(x,y))^\gamma
$

\vspace{0.4em}

$\displaystyle
\mathcal D_{\text{nerf}}
=
\{(I_i,w_i)\}_{i=1}^{N}
$
};

% Third box
\node[align=center,scale=0.55]
at (0.893,0.528)
{
\footnotesize
$\displaystyle
c_{\text{GS}}
=
\sum_{i=1}^{N}
c_i\alpha_i\tau_i
$
};

\node[align=center,scale=0.62]
at (0.75,0.644)
{
\footnotesize
$\displaystyle
\mathcal{D}_{\text{NeRF}}
$
};

\node[align=center,scale=0.62]
at (0.63,0.12)
{
\footnotesize
$\displaystyle
\mathcal{D}_{\text{NeRF}}
$
};

\node[align=center,scale=0.7]
at (0.753,0.283)
{
\footnotesize
$\displaystyle
\mathcal{L}_{\text{nerf}}=\sum w_i \| I_i^\text{GS} - I_i^\text{nerf} \|_2^2
$
};

\node[align=center,scale=0.8]
at (0.915,0.27)
{
\footnotesize
$\mathcal{L} = \lambda(t)\mathcal{L}_{\text{nerf}} +$\\
\footnotesize
$(1-\lambda(t))\mathcal{L}_{\text{real}}$
};

\end{scope}
\end{tikzpicture}

\caption{Overview of the AugSplat pipeline.}
\end{figure*}
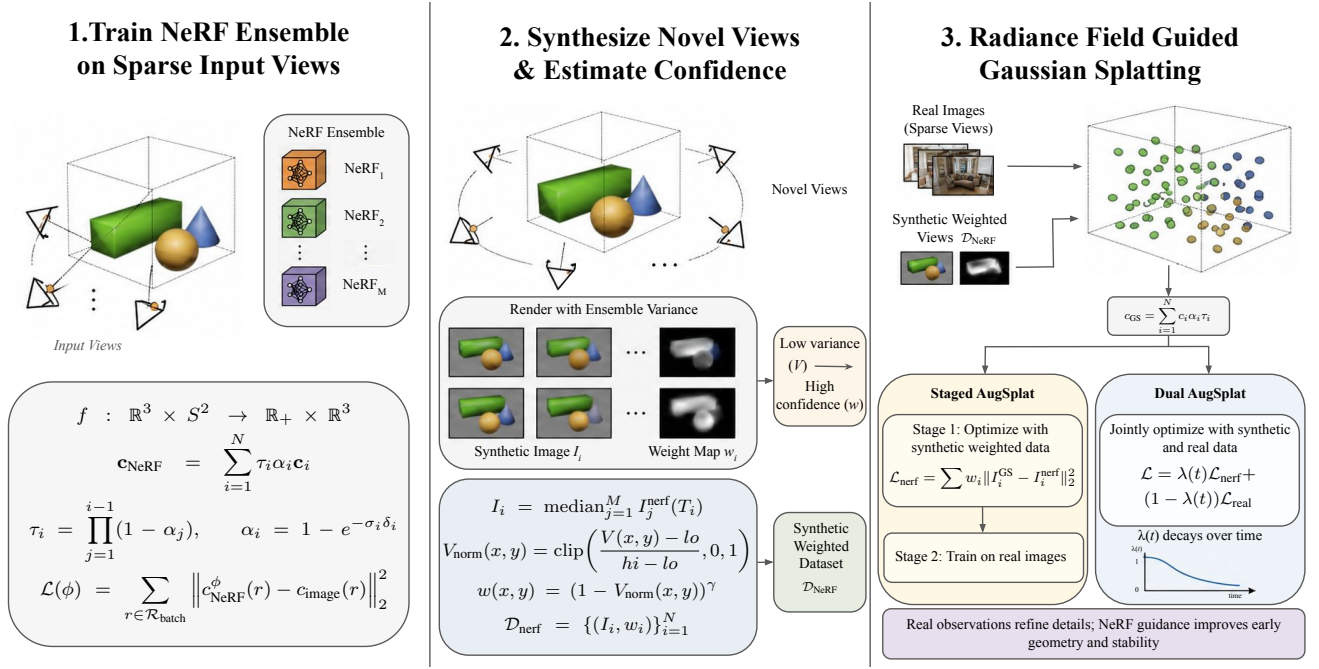

For each synthetic pose, we render one image from each ensemble member and aggregate the predictions using the per-pixel median:
\begin{equation}
    I_i^{\text{nerf}}(\mathbf{p}) =
    \operatorname{median}_{m=1}^{M}
    I^{\text{nerf}}_{i,m}(\mathbf{p}),
\end{equation}
where $\mathbf{p}$ indexes image pixels. This yields a set of synthetic images
$\mathcal{I}_{\text{nerf}} = \{ I_i^{\text{nerf}} \}_{i=1}^{N_{\text{syn}}}$.

To estimate per-pixel confidence, we compute the variance of the ensemble color predictions at each pixel. Let $V_i(\mathbf{p})$ denote the average RGB variance across ensemble members for synthetic view $i$. We normalize this variance using per-image robust lower and upper bounds, computed from the ensemble variance map:
\begin{equation}
    v_{\text{lo},i} = P_5(V_i), 
    \quad
    v_{\text{hi},i} = P_{95}(V_i),
\end{equation}
where $P_5$ and $P_{95}$ denote the 5th and 95th percentiles, respectively. The normalized variance is then given by
\begin{equation}
    V_{\text{norm},i}(\mathbf{p}) =
    \mathrm{clip}\left(
    \frac{V_i(\mathbf{p}) - v_{\text{lo},i}}
    {v_{\text{hi},i} - v_{\text{lo},i}},
    0, 1
    \right).
\end{equation}

The normalized variance is converted into a confidence weight:
\begin{equation}
    w_i(\mathbf{p}) = \left(1 - V_{\text{norm},i}(\mathbf{p})\right)^{\gamma}.
\end{equation}

This yields a weighted synthetic dataset
$\mathcal{D}_{\text{nerf}} =
\{ (I_i^{\text{nerf}}, w_i) \}_{i=1}^{N_{\text{syn}}}$.

Intuitively, the confidence weights encourage the Gaussian model to focus on stable, low-frequency structures in the synthetic views rather than overfitting uncertain high-frequency details produced by the radiance field. This is particularly useful during the early stages of Gaussian optimization, when the primitives are still coarse and the main objective is to recover a reliable geometric layout. As training progresses, supervision from real images can then refine high-frequency appearance details.

\begin{figure}[H]
    \centering
    \includegraphics[width=0.48\columnwidth]{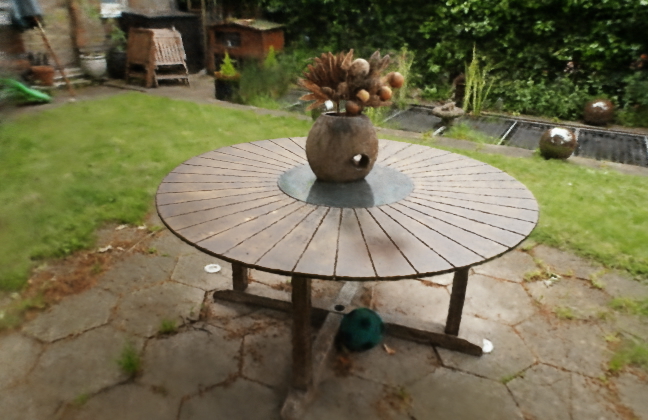}
    \hfill
    \includegraphics[width=0.48\columnwidth]{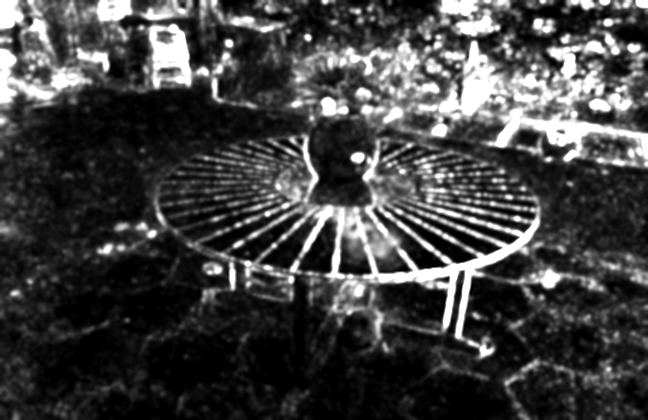}
    \caption{Example NeRF-rendered synthetic view and corresponding confidence map used for weighted supervision.}
    \label{fig:two_images}
\end{figure}

\subsection{Radiance Field-Guided Gaussian Splatting}

\boldparagraph{3D Gaussian Splatting}
3D Gaussian Splatting (3DGS) represents a scene as a set of $G$ anisotropic Gaussian primitives parameterized by position, opacity, color, scale, and rotation. Rendering is performed by rasterizing and alpha-compositing visible Gaussians:
\begin{equation}
    \mathbf{c}_{\text{GS}} = \sum_{g=1}^{G} \mathbf{c}_g \alpha_g \tau_g,
\end{equation}
with transmittance defined analogously to NeRF.

While enabling efficient real-time rendering, 3DGS relies on an explicit set of primitives initialized from reconstructed geometry. Although densification can add new primitives during training, the optimization remains strongly influenced by the initial point cloud and the camera views that provide gradient signal. In sparse-view settings, missing or inaccurate initial geometry can therefore lead to poor local optima.

\boldparagraph{AugSplat}
We propose \textit{AugSplat}, a framework that leverages radiance-field-generated views to improve the early stages of Gaussian Splatting optimization. Our key insight is that nearby synthetic views provide additional geometric and photometric constraints when Gaussian primitives are still coarse and highly deformable.

AugSplat uses two sources of supervision: synthetic views rendered from the radiance-field ensemble and real sparse input images. For synthetic supervision, we sample a minibatch $\mathcal{B}_{\text{syn}}$ uniformly from the weighted synthetic dataset $\mathcal{D}_{\text{nerf}}$ and minimize the per-pixel weighted photometric loss
\begin{equation}
    \mathcal{L}_{\text{nerf}} =
    \sum_{i \in \mathcal{B}_{\text{syn}}}
    \sum_{\mathbf{p}}
    w_i(\mathbf{p})
    \left\|
    I_i^{\text{GS}}(\mathbf{p}) -
    I_i^{\text{nerf}}(\mathbf{p})
    \right\|_2^2 .
\end{equation}
The synthetic images are sampled randomly during training, and the confidence weights reduce the influence of pixels where the radiance-field ensemble is uncertain.

For real-image supervision, each optimization step samples a single real training image $I_i$ uniformly ($i \sim \mathcal{U}$) from the sparse training set. We use the photometric loss
\begin{equation}
    \mathcal{L}_{\text{real}}
    =
    (1 - \lambda)\,\| I_i^{\text{GS}} - I_i\|_2^2
    +
    \lambda \,\mathcal{L}_{\text{SSIM}}(I_i^{\text{GS}}, I_i),
\end{equation}
where $\mathcal{L}_{\text{SSIM}}$ denotes the SSIM-based dissimilarity term. We use the default value of $\lambda=0.2$.

We consider two strategies for combining these supervision sources.

\textbf{Staged AugSplat.}
Staged AugSplat first optimizes Gaussian Splatting using only the synthetic loss $\mathcal{L}_{\text{nerf}}$ for a fixed warm-up period. After this phase, the synthetic supervision is removed and training continues using only the real-image loss $\mathcal{L}_{\text{real}}$. This schedule uses radiance-field views to stabilize the early geometry, while relying on real images for the final appearance refinement.

\textbf{Dual AugSplat.}
Dual AugSplat optimizes Gaussian Splatting using real and synthetic supervision in parallel:
\begin{equation}
    \mathcal{L}(t)
    =
    \lambda_{\text{nerf}}(t)\mathcal{L}_{\text{nerf}}
    +
    \lambda_{\text{real}}(t)\mathcal{L}_{\text{real}}.
\end{equation}
We use an exponentially decaying raw weight for the synthetic NeRF supervision:
\begin{equation}
    \tilde{\lambda}_{\text{nerf}}(t)
    =
    w_{\text{nerf},0}
    \exp\left(-\frac{\ln 4}{D}t\right)
    =
    w_{\text{nerf},0}4^{-t/D},
\end{equation}
where $D$ is the quarter-decay period, i.e., the raw NeRF weight is reduced by a factor of four every $D$ steps. The real-image loss uses a constant raw weight
$\tilde{\lambda}_{\text{real}} = w_{\text{real}}$.
The final loss weights are normalized as
\begin{equation}
    \lambda_{\text{nerf}}(t)
    =
    \frac{
    \tilde{\lambda}_{\text{nerf}}(t)
    }{
    \tilde{\lambda}_{\text{nerf}}(t) + \tilde{\lambda}_{\text{real}}
    },
    \quad
    \lambda_{\text{real}}(t)
    =
    \frac{
    \tilde{\lambda}_{\text{real}}
    }{
    \tilde{\lambda}_{\text{nerf}}(t) + \tilde{\lambda}_{\text{real}}
    }.
\end{equation}
This schedule gives synthetic views greater influence early in training, when Gaussian primitives are still coarse, and gradually shifts the optimization toward real-image supervision for appearance refinement.

\subsection{Implementation Details}

We implement our pipeline using Nerfstudio~\cite{nerfstudio} and gsplat~\cite{ye2025gsplat}. For the radiance-field prior, we use depth-nerfacto, which incorporates depth supervision to improve geometric consistency.

\boldparagraph{NeRF Training}
We train an ensemble of NeRF models using depth-nerfacto, leveraging the estimated poses and depth supervision. The ensemble is used to render synthetic views and compute per-pixel uncertainty estimates.

\boldparagraph{Gaussian Splatting Training}
All Gaussian Splatting variants use the same initialization and base gsplat hyperparameters.

For Staged AugSplat, the initial synthetic warm-up phase uses batches of 20 NeRF-rendered images per step, rendered at $8\times$ lower resolution. After $T_{\text{stage}}=300$ synthetic warm-up steps, training switches to standard real-image supervision with one real image per step.

For Dual AugSplat, each optimization step uses both supervision sources in parallel: one real image at the standard training resolution and a batch of 20 NeRF-rendered images at $8\times$ lower resolution. The synthetic loss is weighted by the exponentially decaying coefficient described above, while the real-image loss uses a constant raw weight.

\boldparagraph{Hyperparameters}
Unless otherwise specified, we use 30 sparse input images per scene, with 26 images used for training and 4 for validation. We train an ensemble of $M=5$ depth-nerfacto models and render $N_{\text{syn}}=200$ synthetic views per scene. For confidence normalization, we use the 5th and 95th percentiles of each synthetic view's variance map as robust lower and upper bounds, and set the confidence exponent to $\gamma=1.5$. For Dual AugSplat, we use raw loss weights $w_{\text{nerf},0}=2$ and $w_{\text{real}}=1$, with the NeRF weight decaying by a factor of four every $D=300$ steps.
The AugSplat hyperparameters were selected pragmatically based on small-scale pilot experiments, dataset size, and training cost. We primarily tuned the relative loss weights and the synthetic warm-up duration, while using fixed values for the number of synthetic views and confidence exponent. We use a short synthetic warm-up period because NeRF-rendered views are most useful early in optimization, when Gaussian primitives are still coarse, while longer synthetic-only training can overfit to artifacts in the radiance-field prior. Similarly, in Dual AugSplat, the NeRF loss is initialized with a larger weight than the real-image loss but is decayed rapidly, encouraging synthetic views to stabilize early geometry without dominating later appearance refinement. The confidence-weighting parameters are chosen to robustly normalize per-image ensemble variance and suppress uncertain synthetic pixels while preserving stable low-frequency structure.
\section{Experiments}
\label{sec:experiments}

\boldparagraph{Dataset}
We evaluate our method on the mip-NeRF 360 dataset~\cite{barron2022mip}, which consists of nine unbounded indoor and outdoor scenes. To simulate sparse-view settings, we subsample 30 images per scene. Of these, 26 images are used for training and 4 images are held out for evaluation.

To ensure that all methods operate under the same sparse-view reconstruction setting, camera poses, depth estimates, and the sparse point cloud are computed from the same selected 30 views. We obtain COLMAP-compatible reconstructions using VGGT~\cite{wang2025vggt}. The resulting poses and depth maps are used to train the depth-supervised radiance-field prior, and the same reconstruction is used consistently across all Gaussian Splatting variants.

\boldparagraph{Baselines}
We compare against standard 3D Gaussian Splatting implemented in the gsplat library.

\boldparagraph{Metrics and Evaluation}
We report PSNR, SSIM, and LPIPS on the held-out evaluation views. Since sparse-view Gaussian Splatting can overfit after reaching its best reconstruction quality, we evaluate models throughout training and report the checkpoint that achieves the lowest average error on the held-out evaluation views.

Following~\cite{barron2021mipnerfmultiscalerepresentationantialiasing}, we use the combined \emph{average error} metric, defined as the geometric mean of MSE, $\sqrt{1-\text{SSIM}}$, and LPIPS:
\begin{equation}
    \text{avg. error} = \left( 10^{-\text{PSNR}/10} \cdot \sqrt{1-\text{SSIM}} \cdot \text{LPIPS} \right)^{1/3}.
\end{equation}

To further analyze training dynamics, we additionally report all methods at the iteration where standard GSplat achieves its lowest average error. This allows us to compare reconstruction quality at an equivalent optimization stage.

\begin{figure*}[t]
    \centering

    % ===================== Kitchen row =====================
    \begin{minipage}[t]{0.48\textwidth}
        \vspace{0pt}
        \centering
        \includegraphics[width=\linewidth]{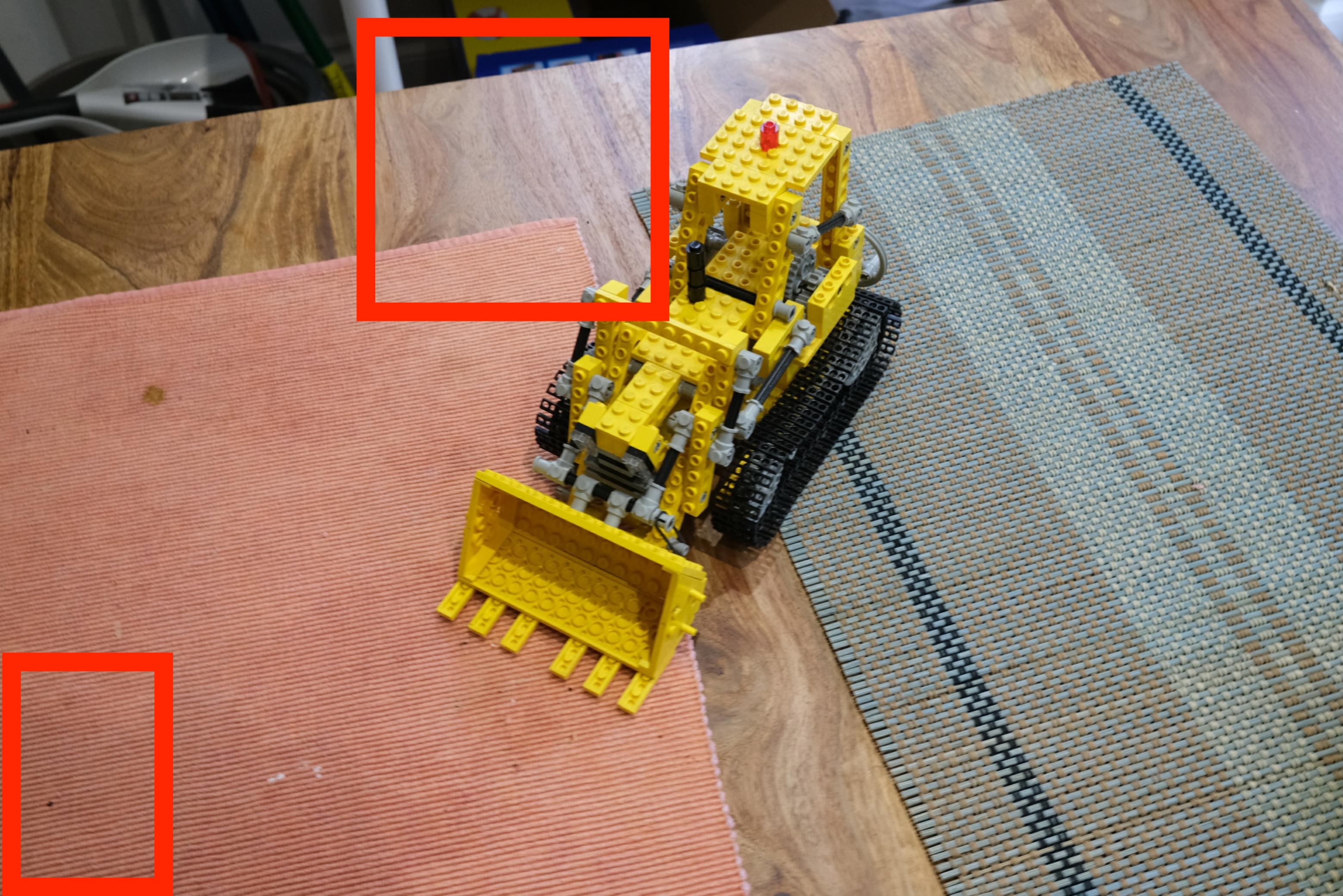}
    \end{minipage}
    \hspace{0.15em}
    \begin{minipage}[t]{0.45\textwidth}
        \vspace{0pt}
        \centering
        \includegraphics[width=0.40\linewidth]{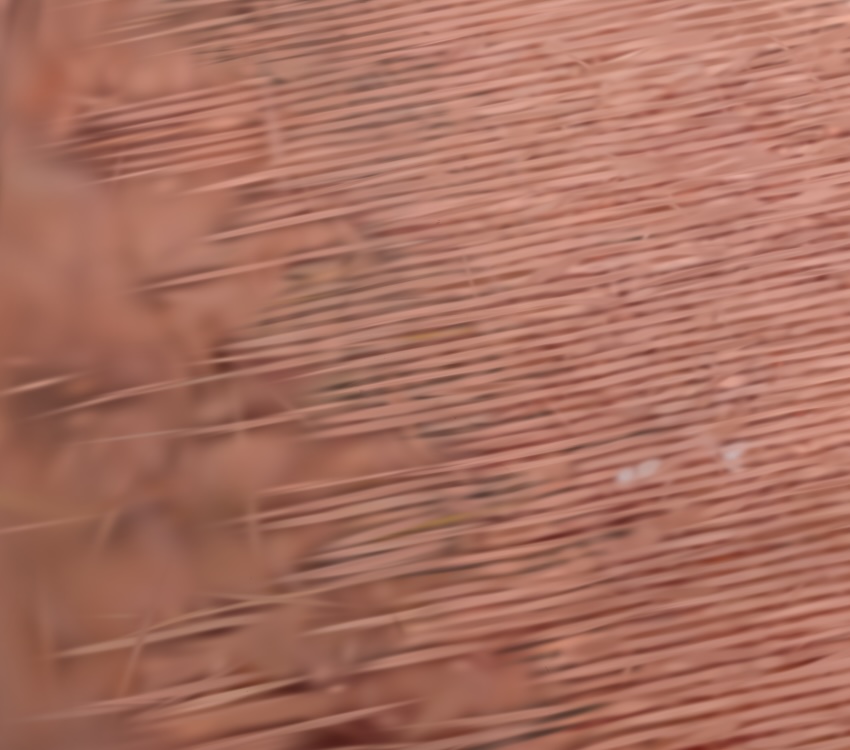}
        \includegraphics[width=0.40\linewidth]{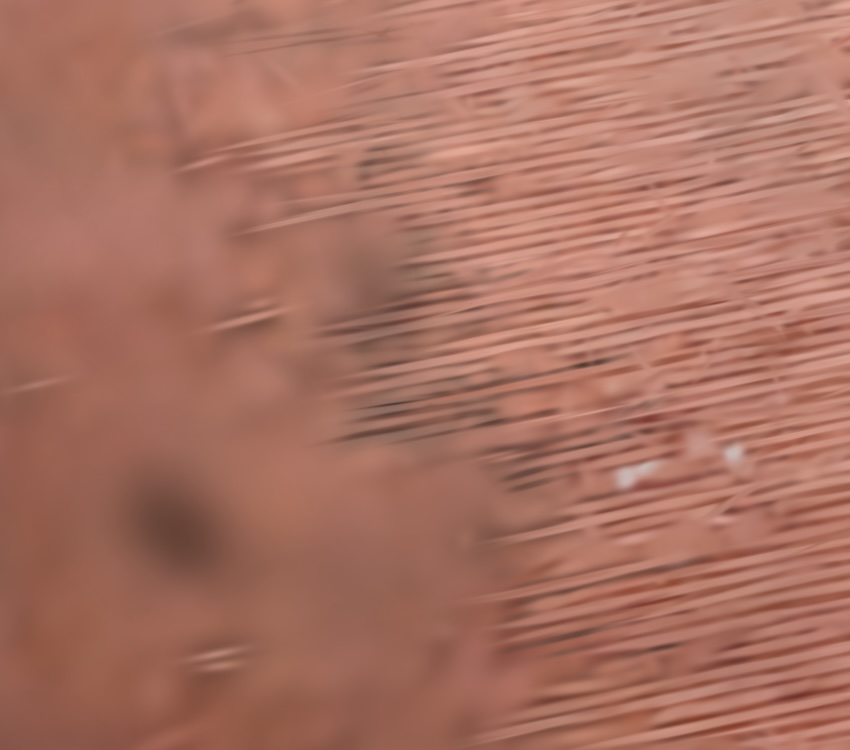}

        \vspace{0.35em}

        \includegraphics[width=0.40\linewidth]{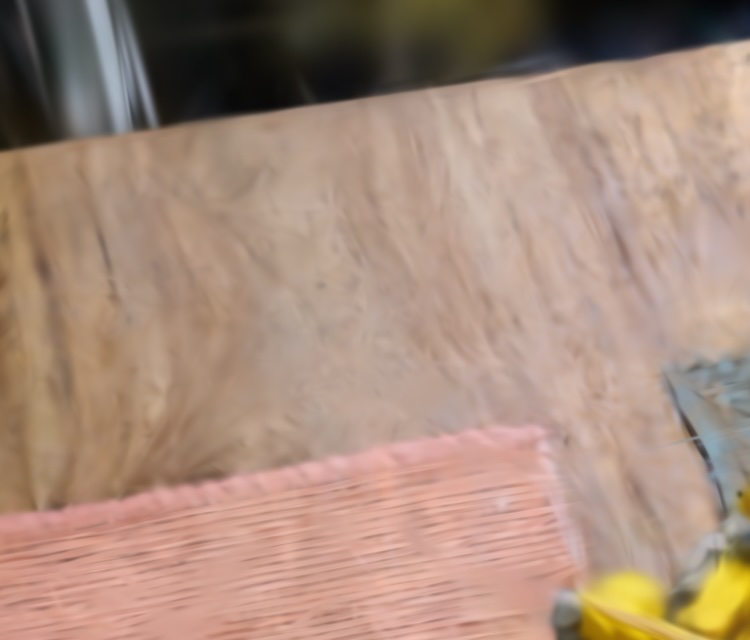}
        \includegraphics[width=0.40\linewidth]{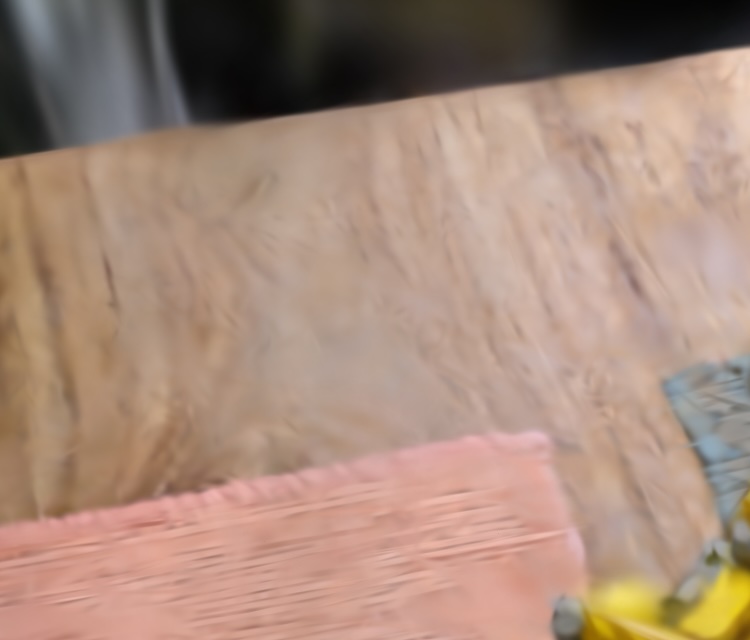}
    \end{minipage}

    \vspace{0.8em}

    % ===================== Garden row =====================
    \begin{minipage}[t]{0.48\textwidth}
        \vspace{0pt}
        \centering
        \includegraphics[width=\linewidth]{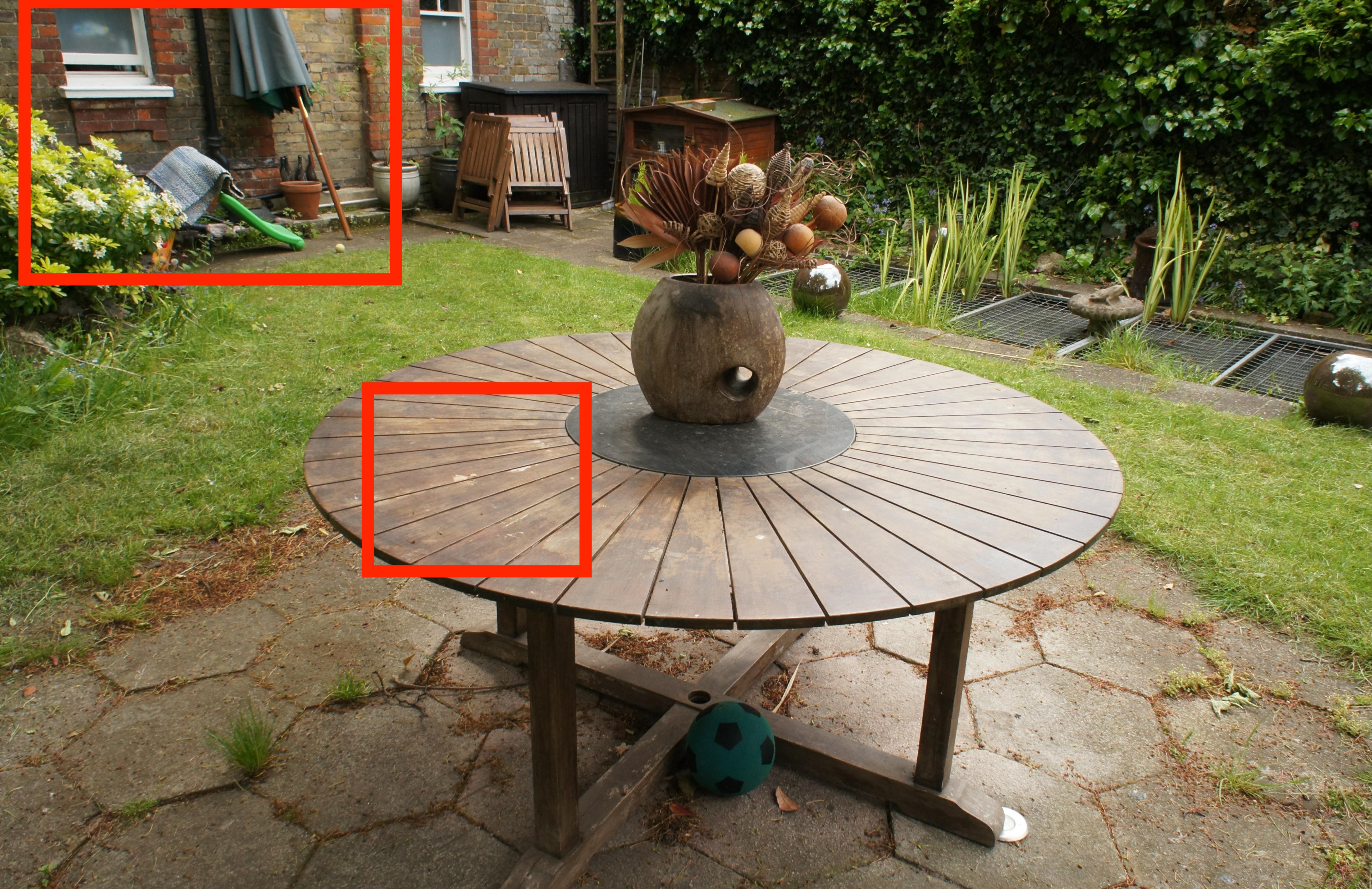}
    \end{minipage}
    \hspace{0.15em}
    \begin{minipage}[t]{0.45\textwidth}
        \vspace{0pt}
        \centering
        \includegraphics[width=0.40\linewidth]{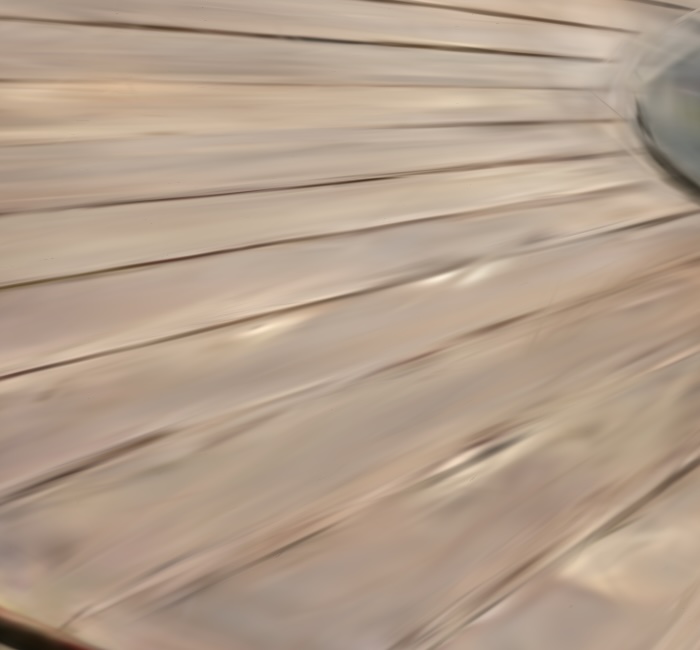}
        \includegraphics[width=0.40\linewidth]{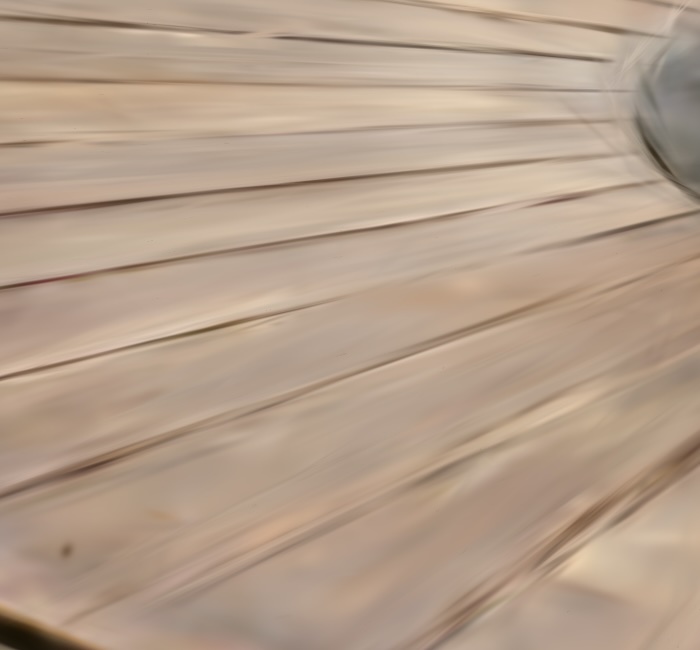}

        \vspace{0.35em}

        \includegraphics[width=0.40\linewidth]{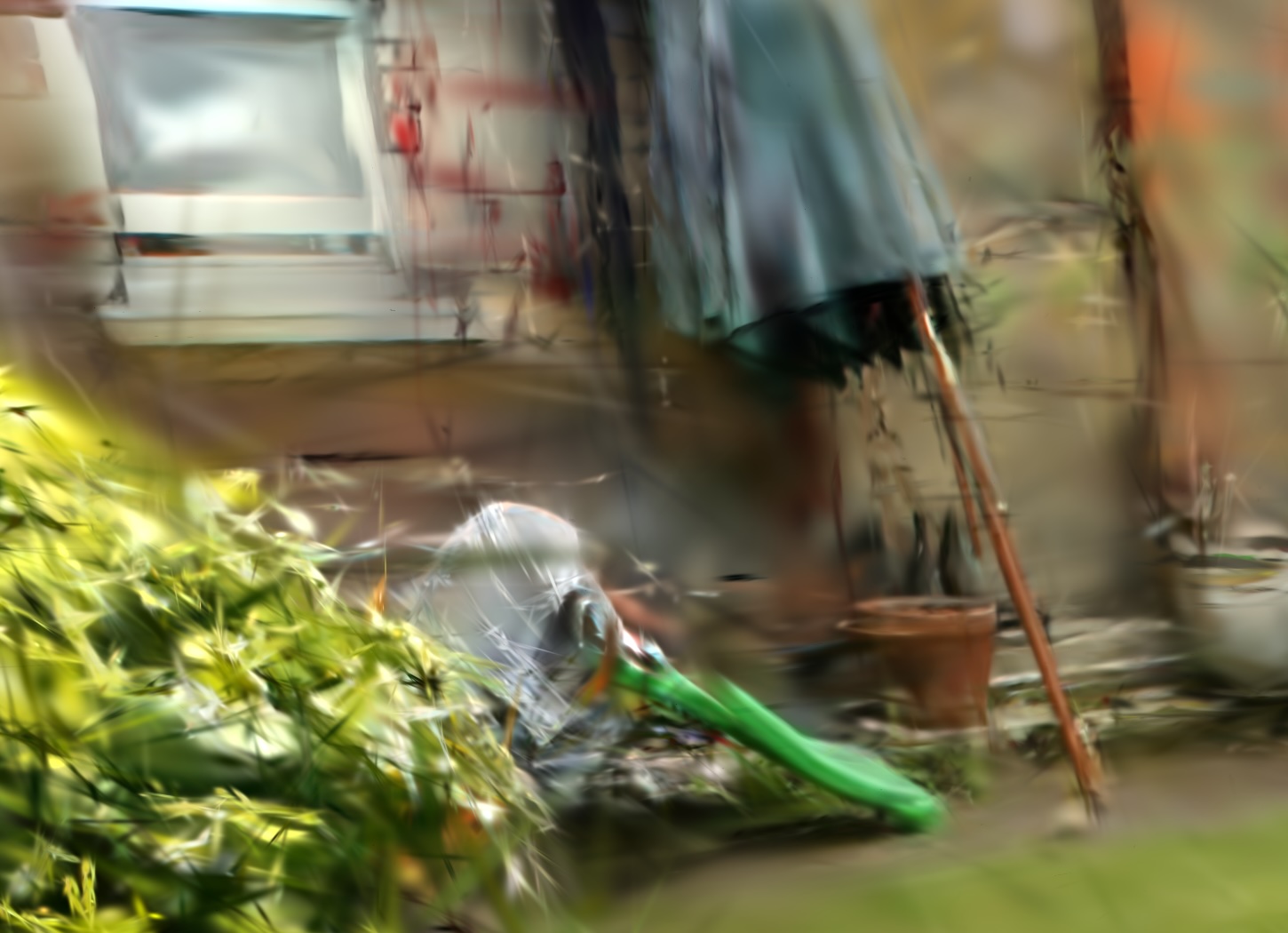}
        \includegraphics[width=0.40\linewidth]{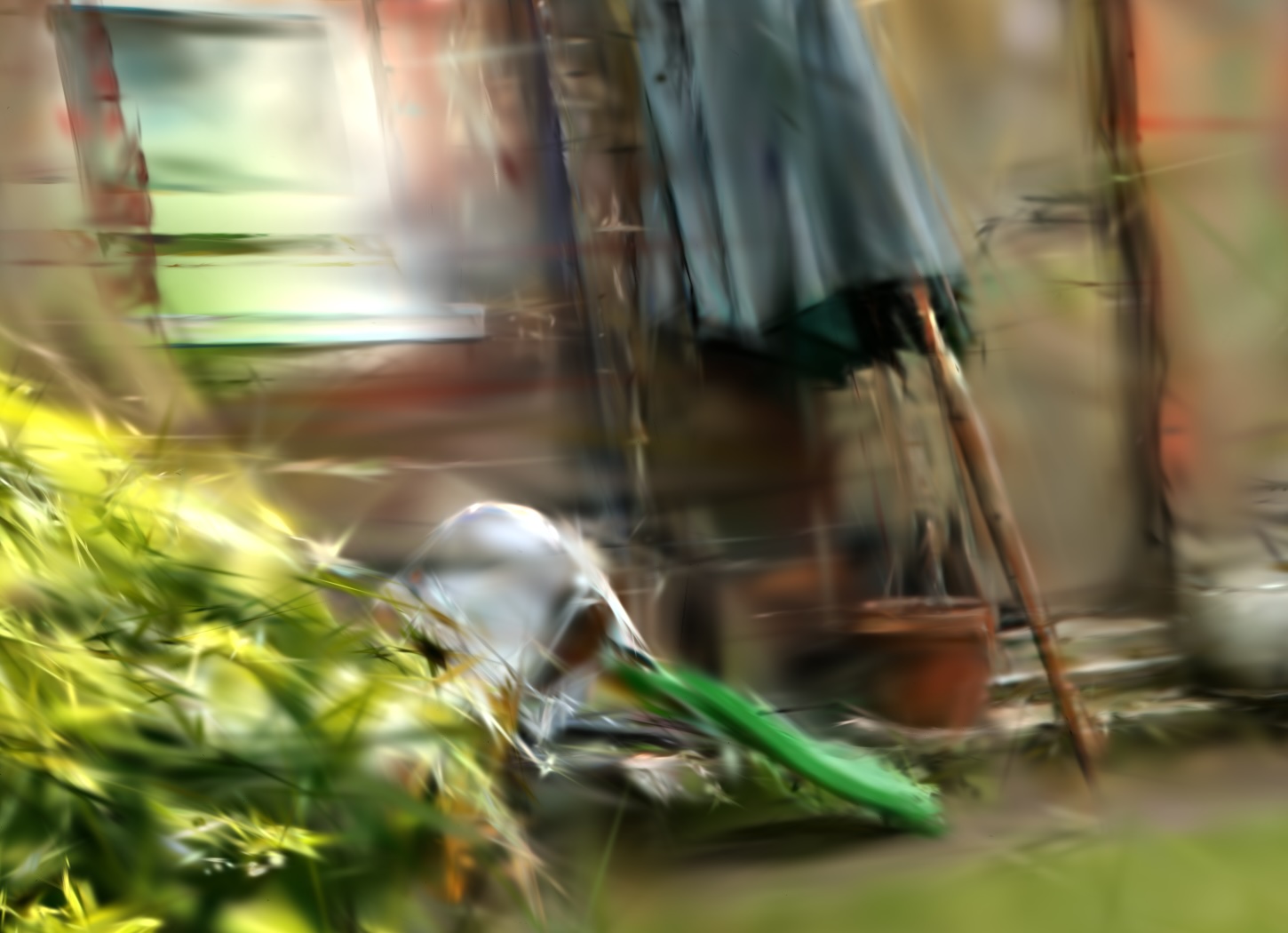}

        \vspace{0.35em}

        % Shared labels for the two detail columns
        \makebox[0.49\linewidth][c]{\small Staged AugSplat}
        \hfill
        \makebox[0.49\linewidth][c]{\small Standard GSplat}
    \end{minipage}

    \caption{
    Visual comparison between Staged AugSplat and standard GSplat on the kitchen and garden scenes. In the kitchen scene, standard GSplat creates a dark dot in the first detailed crop, while Staged AugSplat is more precise and less blurry in reproducing the placemats in the second crop. In the garden scene, standard GSplat is unable to reproduce the white dots in the first detailed crop, while in the second crop Staged AugSplat is more precise in the window and bush, and standard GSplat fails to reproduce the ball at the feet of the parasol.
    }
    \label{fig:qualitative_details}
\end{figure*}

\boldparagraph{Per-Scene Analysis}
The effectiveness of AugSplat depends on scene characteristics. AugSplat yields its largest improvements on challenging scenes such as Bicycle and Flowers, where sparse-view initialization is more likely to miss or misplace geometry. On moderately complex scenes such as Kitchen and Stump, the dual formulation provides consistent gains in average error. On simpler scenes such as Room and Counter, performance remains comparable to the baseline, suggesting that standard Gaussian Splatting is already well constrained in these cases.

Average results are shown in Table~\ref{tab:avg_res}, and per-scene results are reported in Table~\ref{tab:all_results}.

\begin{table*}[t]
\centering
\resizebox{\textwidth}{!}{
\begin{tabular}{l|cccc|cccc|cccc}
\multicolumn{1}{c|}{} 
& \multicolumn{4}{c|}{GSplat} 
& \multicolumn{4}{c|}{Staged AugSplat} 
& \multicolumn{4}{c}{Dual AugSplat} \\
Scene 
& SSIM$\uparrow$ & PSNR$\uparrow$ & LPIPS$\downarrow$ & Avg$\downarrow$
& SSIM$\uparrow$ & PSNR$\uparrow$ & LPIPS$\downarrow$ & Avg$\downarrow$
& SSIM$\uparrow$ & PSNR$\uparrow$ & LPIPS$\downarrow$ & Avg$\downarrow$ \\
\hline

Bicycle 
& 0.584 & 14.21 & 0.831 & 0.273 
& \cellcolor{red!25}0.597 & \cellcolor{red!25}15.32 & \cellcolor{orange!25}0.820 & \cellcolor{red!25}0.252 
& \cellcolor{orange!25}0.594 & \cellcolor{orange!25}15.03 & \cellcolor{red!25}0.816 & \cellcolor{orange!25}0.253 \\

Bonsai 
& 0.735 & 19.84 & 0.532 & 0.142 
& \cellcolor{red!25}0.738 & \cellcolor{red!25}19.95 & \cellcolor{red!25}0.521 & \cellcolor{red!25}0.139 
& \cellcolor{orange!25}0.737 & \cellcolor{orange!25}19.88 & \cellcolor{orange!25}0.523 & \cellcolor{orange!25}0.141 \\

Counter 
& \cellcolor{orange!25}0.753 & 20.31 & \cellcolor{red!25}0.509 & 0.133 
& 0.752 & \cellcolor{orange!25}20.46 & \cellcolor{orange!25}0.519 & \cellcolor{orange!25}0.132 
& \cellcolor{red!25}0.758 & \cellcolor{red!25}20.66 & 0.541 & \cellcolor{red!25}0.132 \\

Flowers 
& 0.445 & 14.36 & 0.909 & 0.292 
& \cellcolor{red!25}0.454 & \cellcolor{red!25}15.23 & \cellcolor{red!25}0.906 & \cellcolor{red!25}0.272 
& \cellcolor{orange!25}0.453 & \cellcolor{orange!25}14.81 & \cellcolor{red!25}0.906 & \cellcolor{orange!25}0.281 \\

Garden 
& \cellcolor{orange!25}0.483 & \cellcolor{orange!25}17.70 & \cellcolor{orange!25}0.612 & \cellcolor{orange!25}0.195 
& 0.482 & 17.57 & \cellcolor{red!25}0.611 & 0.197 
& \cellcolor{red!25}0.487 & \cellcolor{red!25}17.82 & 0.618 & \cellcolor{red!25}0.194 \\

Kitchen 
& \cellcolor{orange!25}0.609 & 19.11 & 0.328 & 0.136 
& 0.608 & \cellcolor{orange!25}19.51 & \cellcolor{red!25}0.317 & \cellcolor{orange!25}0.131 
& \cellcolor{red!25}0.609 & \cellcolor{red!25}19.73 & \cellcolor{orange!25}0.318 & \cellcolor{red!25}0.128 \\

Room 
& \cellcolor{orange!25}0.761 & \cellcolor{orange!25}24.14 & \cellcolor{orange!25}0.421 & \cellcolor{orange!25}0.093 
& 0.756 & 23.90 & \cellcolor{red!25}0.420 & 0.095 
& \cellcolor{red!25}0.763 & \cellcolor{red!25}24.28 & 0.433 & \cellcolor{red!25}0.092 \\

Stump 
& 0.678 & 17.30 & \cellcolor{red!25}0.785 & 0.202 
& \cellcolor{red!25}0.692 & \cellcolor{red!25}18.35 & 0.793 & \cellcolor{red!25}0.186 
& \cellcolor{orange!25}0.689 & \cellcolor{orange!25}18.15 & \cellcolor{orange!25}0.786 & \cellcolor{orange!25}0.189 \\

Treehill 
& 0.547 & 13.98 & \cellcolor{red!25}0.701 & 0.266
& \cellcolor{red!25}0.593 & \cellcolor{red!25}15.19 & 0.777 & \cellcolor{red!25}0.247
& \cellcolor{orange!25}0.577 & \cellcolor{orange!25}14.96 & \cellcolor{orange!25}0.735 & \cellcolor{orange!25}0.248 \\

\hline
Average 
& 0.622 & 17.88 & \cellcolor{red!25}0.625 & 0.192
& \cellcolor{red!25}0.630 & \cellcolor{red!25}18.39 & 0.632 & \cellcolor{red!25}0.183
& \cellcolor{red!25}0.630 & \cellcolor{orange!25}18.37 & \cellcolor{orange!25}0.631 & \cellcolor{orange!25}0.184 \\

\end{tabular}
}
\caption{Per-scene results on mip-NeRF 360. Best results are highlighted in red, second best in orange.}
\label{tab:all_results}
\end{table*}

\begin{table*}[t]
\centering
\begin{subtable}[t]{.48\textwidth}
  \resizebox{\textwidth}{!}{
    \begin{tabular}{l|cccc}
    \multicolumn{1}{c|}{} & SSIM $\uparrow$ & PSNR $\uparrow$ & LPIPS $\downarrow$ & Avg $\downarrow$ \\
    \hline
    GSplat & 0.622 & 17.88 & \textbf{0.625} & 0.192 \\
    Staged AugSplat & \textbf{0.630} & \textbf{18.39} & 0.632 & \textbf{0.183} \\
    Dual AugSplat & \textbf{0.630} & 18.37 & 0.631 & 0.184 \\
    \hline
    \end{tabular}
    }
  \caption{Average performance across all scenes.}
  \label{tab:avg_res}
\end{subtable}
\hfill
\begin{subtable}[t]{.48\textwidth}
  \resizebox{\textwidth}{!}{
    \begin{tabular}{l|cccc}
    \multicolumn{1}{c|}{} & SSIM $\uparrow$ & PSNR $\uparrow$ & LPIPS $\downarrow$ & Avg $\downarrow$ \\
    \hline
    GSplat & 0.622 & 17.88 & 0.625 & 0.192 \\
    Staged AugSplat & 0.624 & 18.23 & 0.626 & \textbf{0.185} \\
    Dual AugSplat & \textbf{0.625} & \textbf{18.25} & \textbf{0.621} & \textbf{0.185} \\
    \hline
    \end{tabular}
    }
  \caption{Performance at GSplat's best iteration.}
  \label{tab:intermediate_steps}
\end{subtable}
\caption{Quantitative results.}
\end{table*}

\boldparagraph{Training Dynamics}

To analyze optimization behavior, we compare all methods at the iteration where standard GSplat achieves its lowest validation average error. As shown in Table~\ref{tab:intermediate_steps}, both AugSplat variants achieve lower average error than the baseline at this equivalent training stage, with Dual AugSplat obtaining the best overall result.

This suggests that the improvements are not simply due to longer training or checkpoint selection, but are already present during early optimization.

\begin{figure}[H]
    \centering
    \includegraphics[width=0.82\linewidth]{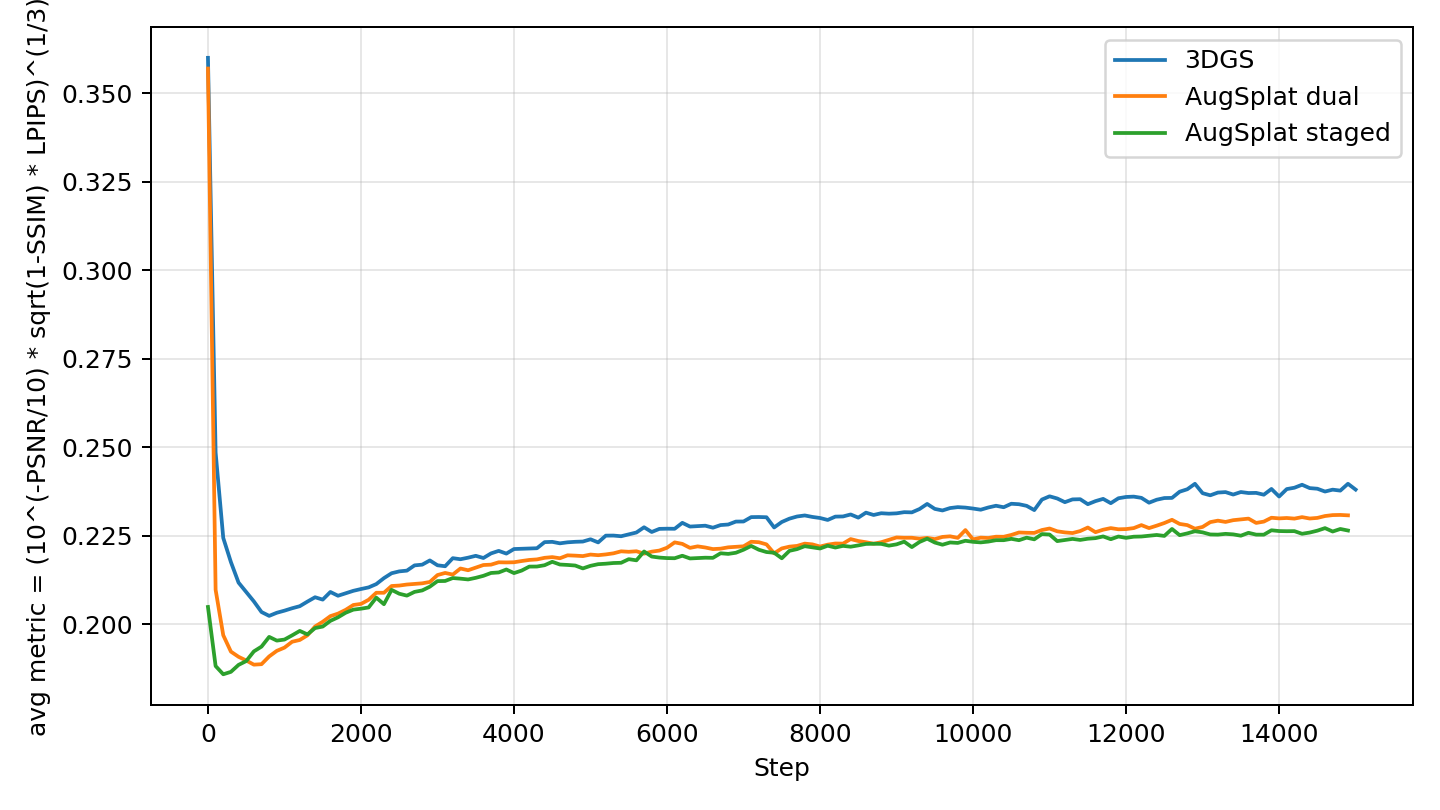}
    \caption{
    Training dynamics on the stump scene. The combined average error reaches its optimum at a compromise point between image-fidelity and perceptual metrics.
    }
    \label{fig:training_dynamics}
\end{figure}

Figure~\ref{fig:training_dynamics} shows representative training curves for sparse-view scenes.
We observe a recurring pattern: PSNR and SSIM often reach their highest values early in training and then gradually decrease, whereas LPIPS continues to improve over later iterations.
As a result, the combined average error is minimized at an intermediate point that balances pixel-level fidelity, structural similarity, and perceptual similarity.

This behavior is consistent with sparse-view overfitting. After the initial optimization stage, the model can continue adapting to the limited training views, improving perceptual similarity according to LPIPS, while losing test-view fidelity as measured by PSNR and SSIM.
AugSplat follows the same general trend, but it typically reaches better PSNR and SSIM during the early optimization phase.
This explains the difference between Table~\ref{tab:intermediate_steps} and Table~\ref{tab:avg_res}: standard 3DGS may obtain its best average error at a later point where LPIPS is lower, but PSNR and SSIM have already degraded, whereas AugSplat often reaches its best average error earlier, when image fidelity is still higher.
AugSplat generally maintains better PSNR and SSIM during the early optimization phase and achieves lower average error at the selected checkpoints.

\begin{figure*}[t]
    \centering
    \includegraphics[width=0.95\textwidth]{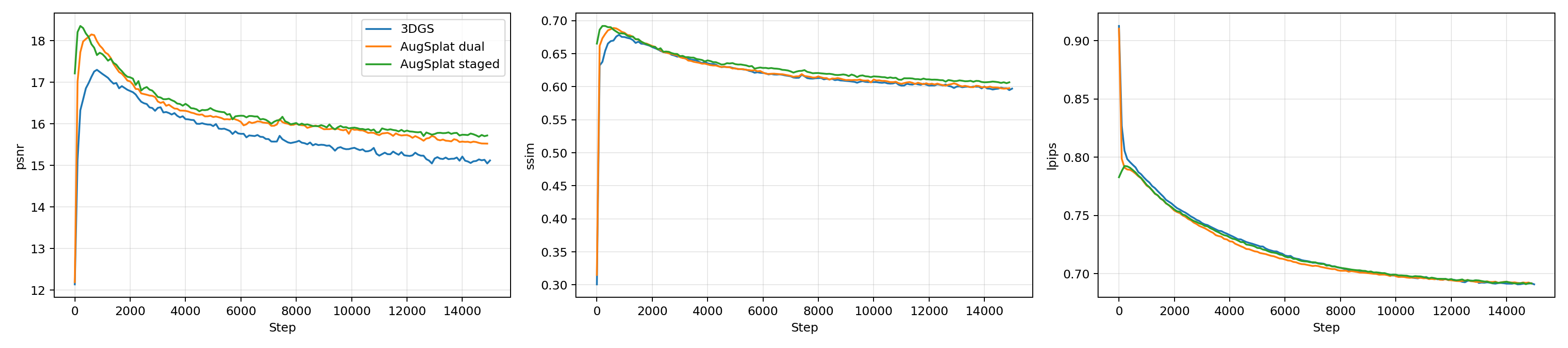}
    \caption{
    Detailed training dynamics for PSNR, SSIM, and LPIPS on a representative sparse-view scene.
    PSNR and SSIM peak early and then gradually decrease, while LPIPS continues to improve over later iterations.
    This illustrates the trade-off captured by the average-error metric.
    }
    \label{fig:training_dynamics_all}
\end{figure*}

\boldparagraph{Summary}
Overall, our results demonstrate that NeRF-based augmentation improves Gaussian Splatting in sparse-view settings. The dual formulation provides a stable alternative that keeps real-image supervision active throughout training, while Staged AugSplat achieves the best average reconstruction quality.

\boldparagraph{Limitations}
While AugSplat improves Gaussian Splatting in many sparse-view scenarios, its effectiveness depends on the quality of the NeRF prior used for augmentation. In highly complex scenes, where the radiance field is less reliable, synthetic views might introduce inaccuracies that negatively affect optimization. 

Furthermore, the staged formulation exhibits higher variance across scenes, occasionally leading to degraded performance when synthetic supervision dominates early training. Although the dual formulation mitigates this issue, it still relies on a careful balance between synthetic and real supervision.

Finally, AugSplat requires training a radiance-field prior in addition to Gaussian Splatting, increasing the overall training complexity.

\section{Conclusion and Future Work}
In this work, we introduced AugSplat, a framework that leverages neural radiance fields to improve Gaussian Splatting in sparse-view settings. By augmenting the training data with NeRF-generated views, we increase the effective supervisory coverage available during training and provide a stronger signal for early-stage optimization.

We proposed two training strategies, Staged AugSplat and Dual AugSplat. In our experiments, Staged AugSplat achieves the best average reconstruction quality, while Dual AugSplat provides a closely performing alternative that combines real and synthetic supervision throughout training. These results suggest that radiance-field-generated views are most useful during the early stages of Gaussian optimization, where they can provide additional geometric and photometric constraints.

Our experiments on the mip-NeRF 360 dataset show that AugSplat improves reconstruction quality in challenging sparse-view scenarios, while remaining competitive in better-constrained scenes. These results indicate that radiance fields can serve as useful auxiliary supervision for explicit Gaussian representations, helping mitigate some limitations of sparse-view Gaussian Splatting.
{
    \small
    \bibliographystyle{ieeenat_fullname}
    \bibliography{main}

@String(CVPR= {IEEE Conf. Comput. Vis. Pattern Recog.})

@String(ICCV= {Int. Conf. Comput. Vis.})

@String(ECCV= {Eur. Conf. Comput. Vis.})

@String(ICLR = {Int. Conf. Learn. Represent.})

@String(CVPR  = {CVPR})

@String(ICCV  = {ICCV})

@String(ECCV  = {ECCV})

@String(ICLR  = {ICLR})

@String(SIGGRAPH  =	{SIGGRAPH})

@inproceedings{qin2024langsplat,
  title={LangSplat: 3D Language Gaussian Splatting},
  author={Qin, Minghan and Li, Wanhua and Zhou, Jiawei and Wang, Haoqian and Pfister, Hanspeter},
  booktitle = CVPR,
  year = {2024}
}

@inproceedings{shi2024language,
  title={Language Embedded 3D Gaussians for Open-Vocabulary Scene Understanding}, 
  author={Jin-Chuan Shi and Miao Wang and Hao-Bin Duan and Shao-Hua Guan},
  booktitle = CVPR,
  year={2024}
}

@article{xie2022neural,
  title   = {Neural Fields in Visual Computing and Beyond},
  author  = {Xie, Yiheng and Takikawa, Towaki and Saito, Shunsuke and
             Litany, Or and Yan, Shiqin and Khan, Numair and
             Tombari, Federico and Tompkin, James and
             Sitzmann, Vincent and Sridhar, Srinath},
  journal = {Computer Graphics Forum},
  volume  = {41},
  number  = {2},
  pages   = {641--676},
  year    = {2022},
  doi     = {10.1111/cgf.14505}
}

@inproceedings{mildenhall2020nerf,
 title={NeRF: Representing Scenes as Neural Radiance Fields for View Synthesis},
 author={Ben Mildenhall and Pratul P. Srinivasan and Matthew Tancik and Jonathan T. Barron and Ravi Ramamoorthi and Ren Ng},
 year={2020},
 booktitle=ECCV,
}

@inproceedings{barron2022mip,
  title={Mip-nerf 360: Unbounded anti-aliased neural radiance fields},
  author={Barron, Jonathan T and Mildenhall, Ben and Verbin, Dor and Srinivasan, Pratul P and Hedman, Peter},
  booktitle=CVPR,
  year={2022}
}

@article{kerbl20233d,
  title={3d gaussian splatting for real-time radiance field rendering},
  author={Kerbl, Bernhard and Kopanas, Georgios and Leimk{\"u}hler, Thomas and Drettakis, George},
  journal=SIGGRAPH,
  year={2023},
}

@inproceedings{mescheder2019occupancynet,
  title = {Occupancy Networks: Learning 3D Reconstruction in Function Space},
  author = {Mescheder, Lars and Oechsle, Michael and Niemeyer, Michael and Nowozin, Sebastian and Geiger, Andreas},
  booktitle = CVPR,
  year = {2019},
}

@inproceedings{park2019deepsdf,
author = {Park, Jeong Joon and Florence, Peter and Straub, Julian and Newcombe, Richard and Lovegrove, Steven},
title = {DeepSDF: Learning Continuous Signed Distance Functions for Shape Representation},
booktitle = CVPR,
year = {2019}
}

@inproceedings{chen2018implicit_decoder,
  title={Learning Implicit Fields for Generative Shape Modeling},
  author={Chen, Zhiqin and Zhang, Hao},
  booktitle=CVPR,
  year={2019}
}

@inproceedings{dhamo2023headgas,
  author = {Dhamo, Helisa and Nie, Yinyu and Moreau, Arthur and Song, Jifei and Shaw, Richard and Zhou, Yiren and P{\'e}rez-Pellitero, Eduardo},
  title = {HeadGaS: Real-Time Animatable Head Avatars via 3D Gaussian Splatting},
  booktitle = {Computer Vision -- ECCV 2024},
  pages = {459--476},
  year = {2024},
  publisher = {Springer Nature Switzerland},
  doi = {10.1007/978-3-031-72627-9_26},
  isbn = {978-3-031-72626-2}
}

@inproceedings{wu20244dgaussians,
      title={4D Gaussian Splatting for Real-Time Dynamic Scene Rendering},
      author={Wu, Guanjun and Yi, Taoran and Fang, Jiemin and Xie, Lingxi and Zhang, Xiaopeng and Wei Wei and Liu, Wenyu and Tian, Qi and Wang, Xinggang},
  booktitle=CVPR,
  year={2024}
}

@inproceedings{matsuki2024gaussian,
  title={Gaussian splatting slam},
  author={Matsuki, Hidenobu and Murai, Riku and Kelly, Paul HJ and Davison, Andrew J},
  booktitle=CVPR,
  year={2024}
}

@inproceedings{li2024sgs,
  title={Sgs-slam: Semantic gaussian splatting for neural dense slam},
  author={Li, Mingrui and Liu, Shuhong and Zhou, Heng and Zhu, Guohao and Cheng, Na and Deng, Tianchen and Wang, Hongyu},
  booktitle={European Conference on Computer Vision},
  pages={163--179},
  year={2024},
  organization={Springer}
}

@inproceedings{yan2023gs,
  author    = {Yan, Chi and Qu, Delin and Xu, Dan and Zhao, Bin and Wang, Zhigang and Wang, Dong and Li, Xuelong},
  title     = {GS-SLAM: Dense Visual SLAM with 3D Gaussian Splatting},
  booktitle = {CVPR},
  year      ={2024},
}

@inproceedings{yi2024gaussiandreamer,
  title={GaussianDreamer: Fast Generation from Text to 3D Gaussians by Bridging 2D and 3D Diffusion Models},
  author={Yi, Taoran and Fang, Jiemin and Wang, Junjie and Wu, Guanjun and Xie, Lingxi and Zhang, Xiaopeng and Liu, Wenyu and Tian, Qi and Wang, Xinggang},
  booktitle=CVPR,
  year={2024}
}

@inproceedings{tang2024dreamgaussian,
  title={DreamGaussian: Generative Gaussian Splatting for Efficient 3D Content Creation},
  author={Tang, Jiaxiang and Ren, Jiawei and Zhou, Hang and Liu, Ziwei and Zeng, Gang},
  booktitle=ICLR,
  year={2024}
}

@article{chung2023luciddreamer,
author = {Chung, Jaeyoung and Lee, Suyoung and Nam, Hyeongjin and Jaerin, Lee and Lee, Kyoung Mu},
year = {2025},
month = {09},
pages = {},
title = {LucidDreamer: Domain-Free Generation of 3D Gaussian Splatting Scenes},
volume = {PP},
journal = {IEEE transactions on visualization and computer graphics},
doi = {10.1109/TVCG.2025.3611489}
}

@inproceedings{moreau2024human,
  title={Human gaussian splatting: Real-time rendering of animatable avatars},
  author={Moreau, Arthur and Song, Jifei and Dhamo, Helisa and Shaw, Richard and Zhou, Yiren and P{\'e}rez-Pellitero, Eduardo},
  booktitle=CVPR,
  year={2024}
}

@inproceedings{qian2024gaussianavatars,
  title={GaussianAvatars: Photorealistic Head Avatars with Rigged 3D Gaussians},
  author={Qian, Shenhan and Kirschstein, Tobias and Schoneveld, Liam and Davoli, Davide and Giebenhain, Simon and Nie\ss{}ner, Matthias},
  booktitle=CVPR,
  year={2024}
}

@inproceedings{zheng2024gpsgaussian,
  title={GPS-Gaussian: Generalizable Pixel-wise 3D Gaussian Splatting for Real-time Human Novel View Synthesis},
  author={Zheng, Shunyuan and Zhou, Boyao and Shao, Ruizhi and Liu, Boning and Zhang, Shengping and Nie, Liqiang and Liu, Yebin},
  booktitle=CVPR,
  year={2024}
}

@inproceedings{Barron2023zipnerf,
  author       = {Jonathan T. Barron and
                  Ben Mildenhall and
                  Dor Verbin and
                  Pratul P. Srinivasan and
                  Peter Hedman},
  title        = {Zip-NeRF: Anti-Aliased Grid-Based Neural Radiance Fields},
  booktitle    = ICCV,
  year         = {2023},
}

@inproceedings{brualla2021wild,
  author       = {Ricardo Martin{-}Brualla and
                  Noha Radwan and
                  Mehdi S. M. Sajjadi and
                  Jonathan T. Barron and
                  Alexey Dosovitskiy and
                  Daniel Duckworth},
  title        = {NeRF in the Wild: Neural Radiance Fields for Unconstrained Photo Collections},
  booktitle    = CVPR,
  year         = {2021},
}

@InProceedings{Sandstrom_2025_CVPR,
    author    = {Sandstr\"om, Erik and Zhang, Ganlin and Tateno, Keisuke and Oechsle, Michael and Niemeyer, Michael and Zhang, Youmin and Patel, Manthan and Van Gool, Luc and Oswald, Martin and Tombari, Federico},
    title     = {Splat-SLAM: Globally Optimized RGB-only SLAM with 3D Gaussians},
    booktitle = {Proceedings of the IEEE/CVF Conference on Computer Vision and Pattern Recognition (CVPR) Workshops},
    month     = {June},
    year      = {2025},
    pages     = {1686-1697}
}

@inproceedings{xu2024fourkfourd,
  author       = {Zhen Xu and
                  Sida Peng and
                  Haotong Lin and
                  Guangzhao He and
                  Jiaming Sun and
                  Yujun Shen and
                  Hujun Bao and
                  Xiaowei Zhou},
  title        = {4K4D: Real-Time 4D View Synthesis at 4K Resolution},
  booktitle    = CVPR,
  year         = {2024},
}

@inproceedings{Niemeyer2021Regnerf,
  author = {Niemeyer, Michael and Barron, Jonathan T. and Mildenhall, Ben and Sajjadi, Mehdi S. M. and Geiger, Andreas and Radwan, Noha},
  title = {RegNeRF: Regularizing Neural Radiance Fields for View Synthesis From Sparse Inputs},
  booktitle = {Proceedings of the IEEE/CVF Conference on Computer Vision and Pattern Recognition (CVPR)},
  month = {June},
  year = {2022},
  pages = {5480--5490},
  doi = {10.1109/CVPR52688.2022.00540}
}

@InProceedings{Schonberger_2016_CVPR,
author = {Schonberger, Johannes L. and Frahm, Jan-Michael},
title = {Structure-From-Motion Revisited},
booktitle = {Proceedings of the IEEE Conference on Computer Vision and Pattern Recognition (CVPR)},
month = {June},
year = {2016}
}

@inproceedings{wang2023visualgeometrygroundeddeep,
  author = {Wang, Jianyuan and Karaev, Nikita and Rupprecht, Christian and Novotny, David},
  title = {VGGSfM: Visual Geometry Grounded Deep Structure From Motion},
  booktitle = {Proceedings of the IEEE/CVF Conference on Computer Vision and Pattern Recognition (CVPR)},
  month = {June},
  year = {2024},
  pages = {21686--21697},
  doi = {10.1109/CVPR52733.2024.02049}
}

@InProceedings{Jain_2021_ICCV,
    author    = {Jain, Ajay and Tancik, Matthew and Abbeel, Pieter},
    title     = {Putting NeRF on a Diet: Semantically Consistent Few-Shot View Synthesis},
    booktitle = {Proceedings of the IEEE/CVF International Conference on Computer Vision (ICCV)},
    month     = {October},
    year      = {2021},
    pages     = {5885-5894}
}

@inproceedings{yu2021pixelnerf,
  author    = {Yu, Alex and Ye, Vickie and Tancik, Matthew and Kanazawa, Angjoo},
  title     = {pixelNeRF: Neural Radiance Fields from One or Few Images},
  booktitle = {Proceedings of the IEEE/CVF Conference on Computer Vision and Pattern Recognition (CVPR)},
  pages     = {4578--4587},
  year      = {2021},
  doi       = {10.1109/CVPR46437.2021.00455}
}

@inproceedings{niemeyer2025radsplat,
  author    = {Niemeyer, Michael and Manhardt, Fabian and Rakotosaona, Marie-Julie and Oechsle, Michael and Duckworth, Daniel and Gosula, Rama and Tateno, Keisuke and Bates, John and Kaeser, Dominik and Tombari, Federico},
  title     = {RadSplat: Radiance Field-Informed Gaussian Splatting for Robust Real-Time Rendering with 900+ FPS},
  booktitle = {2025 International Conference on 3D Vision (3DV)},
  pages     = {134--144},
  year      = {2025},
  doi       = {10.1109/3DV66043.2025.00018}
}

@article{mueller2022instant,
    author = {Thomas M\"uller and Alex Evans and Christoph Schied and Alexander Keller},
    title = {Instant Neural Graphics Primitives with a Multiresolution Hash Encoding},
    journal = {ACM Trans. Graph.},
    issue_date = {July 2022},
    volume = {41},
    number = {4},
    month = jul,
    year = {2022},
    pages = {102:1--102:15},
    articleno = {102},
    numpages = {15},
    url = {https://doi.org/10.1145/3528223.3530127},
    doi = {10.1145/3528223.3530127},
    publisher = {ACM},
    address = {New York, NY, USA}
}

@article{wang2023sparsenerf,
    title={SparseNeRF: Distilling Depth Ranking for Few-shot Novel View Synthesis},
    author={Guangcong Wang and Zhaoxi Chen and Chen Change Loy and Ziwei Liu},
    journal={IEEE/CVF International Conference on Computer Vision (ICCV)},
    year={2023}}

@article{malarz2024gaussiansplattingnerfbasedcolor,
author = {Malarz, Dawid and Smolak-Dyżewska, Weronika and Tabor, Jacek and Tadeja, Sławomir and Spurek, Przemysław},
year = {2025},
month = {12},
pages = {},
title = {Gaussian Splatting with NeRF-based Color and Opacity},
volume = {251},
journal = {Computer Vision and Image Understanding},
doi = {10.1016/j.cviu.2024.104273}
}

@inproceedings{mihajlovic2024splatfieldsneuralgaussiansplats,
   title={SplatFields: Neural Gaussian Splats for Sparse 3D and 4D Reconstruction},
   author={Mihajlovic, Marko and Prokudin, Sergey and Tang, Siyu and Maier, Robert and Bogo, Federica and Tung, Tony and Boyer, Edmond},
   booktitle={European Conference on Computer Vision (ECCV)},
   year={2024},
   organization={Springer}
}

@inproceedings{lu2023scaffoldgsstructured3dgaussians,
  title={Scaffold-gs: Structured 3d gaussians for view-adaptive rendering},
  author={Lu, Tao and Yu, Mulin and Xu, Linning and Xiangli, Yuanbo and Wang, Limin and Lin, Dahua and Dai, Bo},
  booktitle={Proceedings of the IEEE/CVF Conference on Computer Vision and Pattern Recognition},
  pages={20654--20664},
  year={2024}
}

@inproceedings{barron2021mipnerfmultiscalerepresentationantialiasing,
  title={Mip-NeRF: A Multiscale Representation for Anti-Aliasing Neural Radiance Fields},
  author={Barron, Jonathan T. and Mildenhall, Ben and Tancik, Matthew and Hedman, Peter and Martin-Brualla, Ricardo and Srinivasan, Pratul P.},
  booktitle={Proceedings of the IEEE/CVF International Conference on Computer Vision},
  year={2021}
}

@inproceedings{xiong2025sparsegs,
  title={SparseGS: Sparse View Synthesis using 3D Gaussian Splatting},
  author={Xiong, Haolin and Muttukuru, Sairisheek and Xiao, Hanyuan and Upadhyay, Rishi and Chari, Pradyumna and Zhao, Yajie and Kadambi, Achuta},
  booktitle={Proceedings of the International Conference on 3D Vision (3DV)},
  year={2025}
}

@InProceedings{Li_2024_CVPR,
    author    = {Li, Jiahe and Zhang, Jiawei and Bai, Xiao and Zheng, Jin and Ning, Xin and Zhou, Jun and Gu, Lin},
    title     = {DNGaussian: Optimizing Sparse-View 3D Gaussian Radiance Fields with Global-Local Depth Normalization},
    booktitle = {Proceedings of the IEEE/CVF Conference on Computer Vision and Pattern Recognition (CVPR)},
    month     = {June},
    year      = {2024},
    pages     = {20775-20785}
}

@inproceedings{nerfstudio,
	title        = {Nerfstudio: A Modular Framework for Neural Radiance Field Development},
	author       = {
		Tancik, Matthew and Weber, Ethan and Ng, Evonne and Li, Ruilong and Yi, Brent
		and Kerr, Justin and Wang, Terrance and Kristoffersen, Alexander and Austin,
		Jake and Salahi, Kamyar and Ahuja, Abhik and McAllister, David and Kanazawa,
		Angjoo
	},
	year         = 2023,
	booktitle    = {ACM SIGGRAPH 2023 Conference Proceedings},
	series       = {SIGGRAPH '23}
}

@article{ye2025gsplat,
  title={gsplat: An open-source library for Gaussian splatting},
  author={Ye, Vickie and Li, Ruilong and Kerr, Justin and Turkulainen, Matias and Yi, Brent and Pan, Zhuoyang and Seiskari, Otto and Ye, Jianbo and Hu, Jeffrey and Tancik, Matthew and Angjoo Kanazawa},
  journal={Journal of Machine Learning Research},
  volume={26},
  number={34},
  pages={1--17},
  year={2025}
}

@inproceedings{wang2025vggt,
  title={VGGT: Visual Geometry Grounded Transformer},
  author={Wang, Jianyuan and Chen, Minghao and Karaev, Nikita and Vedaldi, Andrea and Rupprecht, Christian and Novotny, David},
  booktitle={Proceedings of the IEEE/CVF Conference on Computer Vision and Pattern Recognition},
  year={2025}
}
}

\end{document}